\pdfoutput=1

\documentclass[11pt]{article}

\usepackage[final]{acl}

\usepackage{times}
\usepackage{latexsym}

\usepackage[T1]{fontenc}

\usepackage[utf8]{inputenc}

\usepackage{microtype}

\usepackage{inconsolata}

\usepackage{graphicx}
\usepackage{langsci-gb4e}
\usepackage{multirow}
\usepackage{booktabs} 
\usepackage{float}
\title{Humans and transformer LMs: Abstraction drives language learning}

\author{Jasper Jian \and Christopher D. Manning\\
  Stanford University \\
  \texttt{jjian@stanford.edu}, 
  \texttt{manning@cs.stanford.edu} \\}

\begin{document}
\maketitle
\begin{abstract}
\textit{Categorization} is a core component of human linguistic competence. We investigate how a transformer-based language model (LM) learns linguistic categories by comparing its behaviour over the course of training to behaviours which characterize \textbf{abstract feature}--based and \textbf{concrete exemplar}--based accounts of human language acquisition. We investigate how lexical semantic and syntactic categories emerge using novel divergence-based metrics that track learning trajectories using next-token distributions. In experiments with GPT-2 small, we find that (i) when a construction is learned, abstract class-level behaviour is evident at earlier steps than lexical item--specific behaviour, and (ii) that different linguistic behaviours emerge abruptly in sequence at different points in training, revealing that abstraction plays a key role in how LMs learn. This result informs the models of human language acquisition that LMs may serve as an existence proof for.
\end{abstract}

\section{Introduction}
Linguistic constructions involve generalizations over classes of words, such as ditransitive or sentential complement-taking verbs, rather than idiosyncratic patterns applying to individual lexical items. Language models (LMs) must learn these categories in order to systematically understand and generate text \citep{pmlr-v80-lake18a, kim-linzen-2020-cogs}. Indeed, work has shown that LMs by the end of training do learn to adeptly categorize. For example, subject-verb agreement \citep{marvin-linzen-2018-targeted, goldberg2019assessing, hao-linzen-2023-verb} requires categorizing singular \textit{vs.}\ plural nouns. This raises the question: what are the \textit{processes} LMs use to learn these generalizations?

\begin{figure}
    \centering
    \includegraphics[width=1\linewidth]{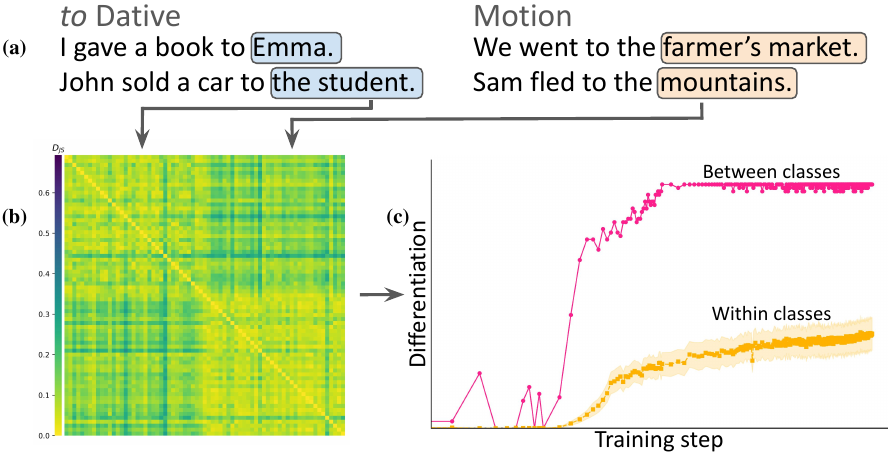}
    \caption{(a) Categories, like verb classes determined by argument structure, are pervasive in language: \textit{to} Dative verbs take recipient arguments, whereas Motion verbs take goal locations. (b) Divergence metrics  compare LM prediction distributions conditioned on different categories. (c) Tracking divergences over training reveals that categories like verb classes are differentiated early by LMs, before gradual item-level learning.}
    \label{fig:verb-class-sentences}
\end{figure}

While interpretability work has begun to decode how trained transformers work, less work has examined their acquisition of knowledge during training \citep{liu-etal-2021-probing-across, choshen-etal-2022-grammar} and even less has compared transformer LM learning with human language learning \citep{chang-bergen-2022-word, evanson-etal-2023-language}. We draw on theories of human language acquisition to tackle the latter question, investigating what hypothesized processes of category formation LM learning corresponds to. We contrast the \textbf{abstraction}-first account and the \textbf{exemplar}-first account which hypothesize that learning privileges (a) linking the input data to abstract featural representations (e.g., \citealp{Pinker1989learnability, Gleitman1990structural, Fisheretal2020developmental}), and (b) storing linguistic examples observed in the input (e.g., \citealp{tomasello1992first, tomasello2003construction, ambridgelieven2015constructivist}), respectively.

We compare the behaviour of GPT-2 over the course of autoregressive pretraining to predictions made by these two accounts, targeting phenomena LMs have been shown to learn. \textbf{Experiment 1} studies how arguments predicted for individual verbs and verb classes change over training (Figure \ref{fig:verb-class-sentences}; \citealp{thrush-etal-2020-investigating, wilson-etal-2023-abstract}) and \textbf{Experiment 2} investigates syntactic subcategorization and non-local dependencies (\citealp{wilcox-etal-2019-hierarchical, wilcox2024using, warstadt-etal-2020-blimp-benchmark}). We find that the abstraction-first account better corresponds to GPT-2 learning across all phenomena tested: linguistic constructions are learned for classes of words at once and individual phenomena are learned sequentially. Furthermore, the ordering of learning onsets corresponds to expected linguistic development in children.

While our experiments involve a single model and do not impose humanlike constraints on training data, our investigation contributes to growing debates around what theories of human linguistic cognition standardly trained autoregressive LMs provide support for \citep{portelance2024roles}. These findings suggest that LMs have a bias towards forming abstractions. Nevertheless, LMs differ from influential abstraction-first accounts of human learning proposing that linguistic categories are innate: LMs learn categories distributionally.

\section{Background}

\subsection{Abstraction-first and exemplar-first learning in children}
How argument structure is learned has motivated two views on the processes underlying human language acquisition:\ the {abstraction-first} account, claiming that children learn structured representations early, and the {exemplar-first} account, claiming that children first memorize linguistic input.

The \textbf{abstraction-first} account is supported by findings like young children's ability to comprehend sentences containing novel verbs by relying on the syntactic structure of the sentence alone \citep[e.g.,][]{naigles1990children} and their incorrect, but meaningful, overextension of syntactic constructions to verbs (e.g., \textit{`you can't happy me'}; \citealp{bowerman1982evaluating}). Abstraction-first accounts explain this as a product of an inductive bias for learning general, abstract properties of language that extend beyond individual words, such as how the syntactic positions of arguments correspond to meanings \citep{landau1985language, Pinker1989learnability, Gleitman1990structural}. Children then use abstract knowledge like this to aid in learning new verbs and to continue refining their representations of previously seen verbs.

The \textbf{exemplar-first} account is supported by findings that young children often only use verbs with nouns \citep{tomasello1992first, tomasello2003construction} and structures \citep{lieven1997lexically} they have they have observed occurring together, as well as evidence that even basic linguistic properties like word order are learned slowly over time \citep{AbbotSmith2001Order}. Exemplar-first accounts propose that these effects arise because children learn by memorizing observed instances of language (`exemplars') resulting in \textit{lexical item--specific} knowledge. Abstractions emerge gradually via repeated exposure and retention, or analogy over stored exemplars \citep{AbbotSmithTomasello2006exemplar, bybee2006, Ambridge2020Against}.

\begin{table*}
    \centering
    \resizebox{2\columnwidth}{!}{
    \begin{tabular}{cccc}
    \hline
       \textsc{Class}  & \textsc{Example Verbs} & \textsc{Example Sentences} & \textsc{Total} \\
    \hline
    (i) \textit{to} Dative & \textit{give}, \textit{sell}, \textit{grant} (\textit{n}=35) & \textit{Chipotle \textbf{gave} away free burritos to the \_\_} & 1841 \\
    (ii) Motion & \textit{go}, \textit{walk}, \textit{flee} (\textit{n}=36) & \textit{Toby accordingly \textbf{goes} to the \_\_} & 1942 \\
    (iii) Reciprocal & \textit{speak}, \textit{meet}, \textit{talk} (\textit{n}=16) & \textit{Cherry recalled Orr had {refused} to \textbf{speak} with the \_\_} & 465 \\
    (iv) \textit{spray-load} & \textit{spray}, \textit{load}, \textit{smear} (\textit{n}=16) & \textit{Ray and Devon then \textbf{sprayed} the table with the \_\_} & 404 \\
    \hline
    \end{tabular}
    }
    \caption{Example verbs and sentence prefixes from the four verb classes used in the experiments.}
    \label{tab:example-data}
\end{table*}

What occurs over the course of learning differentiates the two accounts, in particular, when class-based behaviour \textit{vs.}\ lexical-item specific behaviour emerge. Our experiments exploit this difference to probe whether the learning biases of standard transformer LMs can be characterized as being abstraction-first or exemplar-first.

\subsection{Related work on language models}
Work in NLP and ML focused on downstream task performance or privacy and copyright concerns has discussed the extent to which LMs (and DNNs) implement generalization and memorization mechanisms \citep[a.o.]{arpit2017closer, carlini2019secret, carlini2023quantifying, tirumala2022memorization, huang-etal-2024-demystifying}. We investigate these mechanisms further in the context of how LMs learn linguistic phenomena motivated by hypotheses about human language acquisition.

Previous linguistic work has shown that LMs learn phenomena that abstract over specific lexical items, looking at fully-trained LMs \citep{gulordava-etal-2018-colorless, newman-etal-2021-refining, lasri-etal-2022-bert, papadimitriou-etal-2022-classifying-grammatical, jian-reddy-2023-syntactic} or finetuning fully-trained LMs \citep{kim-smolensky-2021-testing, wilson-etal-2023-abstract, misra-and-kim-2023-catabs}. We suggest that learning \textit{trajectories} can better disambiguate the mechanisms and biases active in LMs \citep[e.g.,][]{kallini-etal-2024-mission, chen2024sudden, chang-etal-2024-characterizing, dankers-titov-2024-generalisation, antonello2024evidence} that may explain the stability of linguistic learning across models and datasets \citep{liu-etal-2021-probing-across, choshen-etal-2022-grammar, evanson-etal-2023-language} and their ability to generalize even to structures absent in their training data \citep{Potts2024Characterizing, misra-mahowald-2024-language, patil-etal-2024-filtered}. Closest to our work, \citet{misra-and-kim-2023-catabs} compare exemplar and abstraction theories in a nonce word learning setting and \citet{yang-etal-2025-transformer} in a transformer-based speech model, both settling on a middle ground conclusion of `abstractions encoded by exemplars.' However, they only evaluate fully-trained models -- \citet{huang-etal-2024-demystifying} show that memorization (i.e., exemplar-storage) abilities change over the course of LM training, emphasizing the need to evaluate behaviours during training to provide a fuller picture of LM mechanisms, as we do here.

\section{Experiment 1: What drives argument structure learning in LMs?}
\subsection{Data and Models}\label{sec:data-and-models}
\paragraph{Empirical domain} Argument structure classes allow us to compare item-specific and class-general effects during learning. We measure the arguments predicted for each verb as a proxy for verb categorization. Four argument structure classes were used which each have arguments introduced by prepositions: (i) \textbf{\textit{to}-dative} verbs, (ii) verbs of \textbf{motion}, (iii) \textbf{reciprocal} verbs, and (iv) \textbf{\textit{spray}-\textit{load}} verbs (see Table \ref{tab:example-data}). Classes (i) and (ii) use the preposition \textit{to}, and (iii) and (iv) use \textit{with}. The arguments in the post-preposition slot form semantic classes \citep{Pinker1989learnability, Dowty2003dual}: (i) recipients, (ii) goal locations, (iii) reciprocal objects, and (iv) substances. LMs must abstract over the specific nouns that they observe with each verb to learn the broad class of nouns that can occur in a given the argument slot.

\paragraph{Trajectories} Early changes to predictions involving learning broad categories of nouns for the argument slots across a group of verbs would provide evidence of an abstraction-first mechanism. Early changes involving idiosyncratic learning for individual verbs, with class-like behaviour emerging later, would evidence an exemplar-first mechanism.\vspace{-\baselineskip}
\begin{figure}
    \centering
    \includegraphics[width=0.9\linewidth]{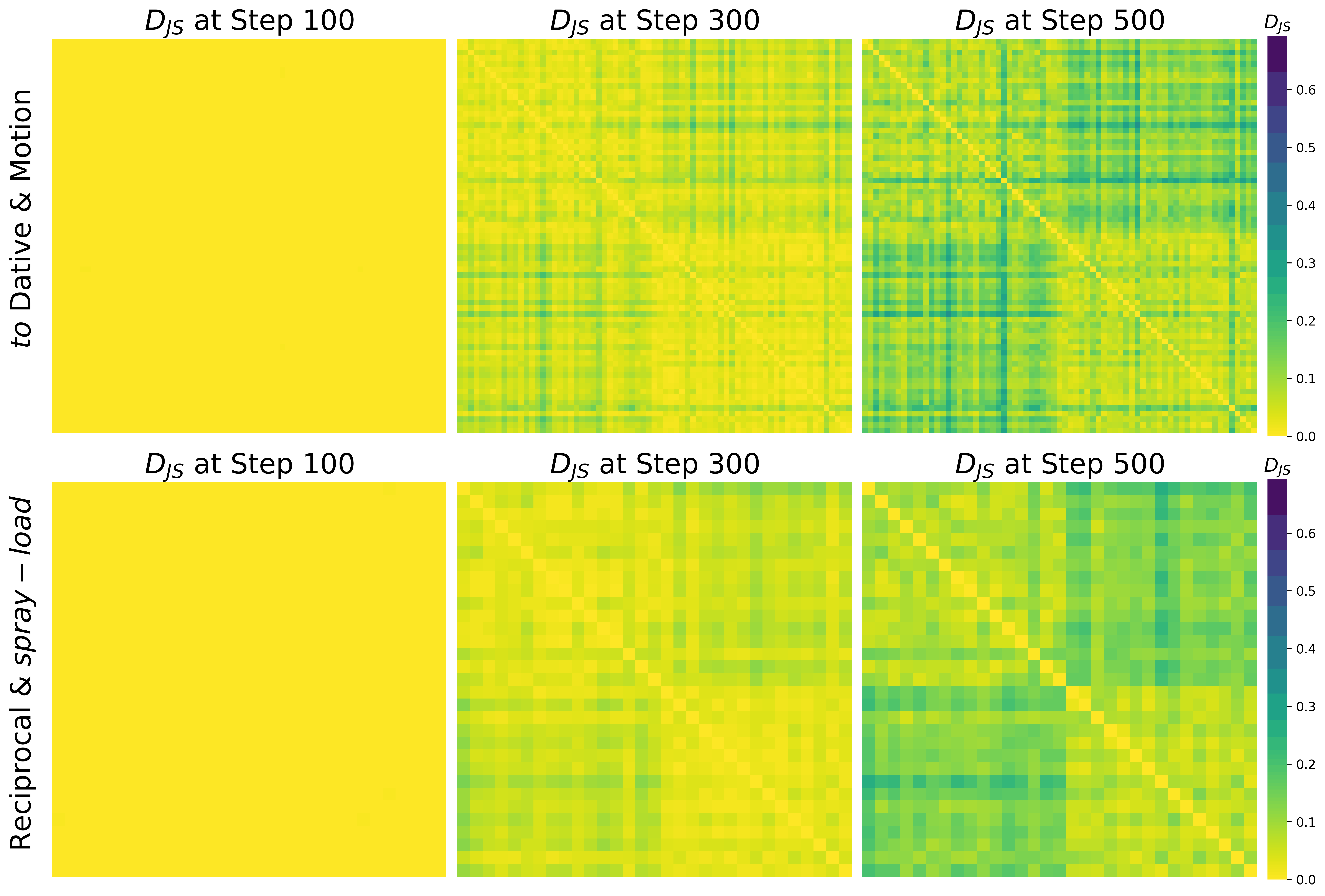}
    \caption{Pairwise $D_{JS}$ between verbs in two classes at three early steps in training. Each row and column represents a single verb. Differentiation into verb classes is evident when any change has occurred -- between-class quadrants have higher $D_{JS}$ (top-right, bottom-left) than in-class quadrants (top-left, bottom-right).}
    \label{fig:jsd-class-plots}
\end{figure}

\paragraph{Dataset} We track the arguments that are predicted by LMs in the target position by taking next-token predictions at the underlined position in sentences containing the structure in (\ref{ex:sample-frame}). Using these distributions, we can track when predictions begin to change when comparing any two verbs (evidence of item-specific learning), and when predictions are more similar for verbs that belong to the same class than for verbs that belong to a separate comparison class (evidence of class-general learning). We compare between pairs of verb classes that share a preposition -- (i) \& (ii), (iii) \& (iv) -- ensuring that differences in argument preferences are the result of verb identity, rather than the identity of the preposition itself.

\noindent \begin{exe}
    \ex ... \textsc{verb} ... \textsc{preposition} \textit{the} \_\_ \label{ex:sample-frame}
\end{exe}

Our dataset contains natural sentences that feature the target verbs in the desired syntactic frame filtered from the WikiText-103 dataset \citep{merity2016pointer}. To find target structures, sentences were filtered with regular expressions and Stanza dependency parses  \citep{qi2020stanza}, before being manually annotated by the authors to account for parser errors and non-target senses of the chosen verbs.

\paragraph{Models} We test for these effects over the course of standard pretraining using training checkpoints of GPT-2 models \citep{radford2019language} released by \citet{karamcheti2021mistral}. Each model was autoregressively trained on the OpenWebText corpus \citep{Gokaslan2019OpenWeb}. We track behaviour over 450 checkpoints (400K steps). Models are trained with a batch size of 32. We choose to use GPT-2 small models (decoder-only transformers; 12 layers and 12 attention heads per layer) given that smaller models have been shown to better predict human linguistic behaviour \citep{oh-schuler-2023-surprisal}. These models have been shown to perform well on standard linguistic benchmarks (e.g., \citealp{warstadt-etal-2020-blimp-benchmark}) and have previously been used to evaluate inductive biases of transformer LMs \citep{kallini-etal-2024-mission}. We report results for a single random seed for clarity, though all described behaviour was confirmed across three runs with different seeds.

\subsection{Paradigm and Metrics}\label{sec:paradigm}
\paragraph{Paradigm} For any verb, we collect next-token distributions across different prefixes. Since we are interested in comparing argument structure predictions at the verb level, we derive a single estimate associated with each verb from the individual distributions obtained from each prefix. For each verb $v$ this is defined as the arithmetic mean of the conditional next token distributions over the set $S_v$ of $N$ prefixes containing verb \textit{v}:\footnote{This is equivalent to an estimate of \( P(x | v) \), obtained by marginalizing over contexts \( s_i \) using samples from \( P(x | S_v, v) \) under the assumption that \( P(S_v | v) \) is uniform.}

\begin{center}
    $P_v(x) = \frac{1}{N} \sum_{i=1}^{N}P(x | s_i), s_i \in S_v$.
\end{center}
We assume that different inflections of a single verb (tense, aspect, and agreement) do not make a significant contribution to argument predictions.

We compute pairwise similarity scores between verbs by taking the Jensen–Shannon divergence, $D_{JS}$, between the distributions defined above \citep{Lin1991Divergence}. This allows us to compute similarity without predefining a set of relevant nouns. Divergence metrics like $D_{JS}$ provide natural interpretations of learning \textit{onsets}, as opposed to metrics like accuracy. Consider two linguistic variables \textit{a} and \textit{b}, each contained within prefixes $s_a$ and $s_b$, which should license different continuations, e.g., \textit{a} is a \textit{to} Dative verb and \textit{b} is a Motion verb. We obtain two next-token distributions from the LM
\begin{center}
    $A=P(x|s_a)$, $B=P(x|s_b)$.
\end{center} 
If $D_{JS}(A||B)=0$, then we know that the LM must not have learned to distinguish between variables \textit{a} and \textit{b} because the prediction distributions are maximally similar. Non-zero $D_{JS}(A||B)$ is the necessary condition for a LM to have begun to distinguish variables correctly during training.\footnote{Smoothing is not required as next-token distributions under autoregressive LMs do not contain true zeroes.}

Previous work has used accuracy scores derived from comparing log-probabilities of sequences, e.g., targeted syntactic evaluation \citep{marvin-linzen-2018-targeted, warstadt-etal-2020-blimp-benchmark}, on linguistic benchmarks to measure onsets of learning \citep{choshen-etal-2022-grammar, evanson-etal-2023-language, chen2024sudden}. However, accuracy metrics may be artificially steep \citep{schaeffer2023are} and it is unclear how to determine the onset of learning if, e.g., baseline accuracy decreases before increasing. The divergence metrics we use are smooth, i.e., no concerns with artificial steepness, and lower-bounded by zero, so increases can be directly interpreted as onsets of behavioural change.

\begin{figure}
    \centering
    \includegraphics[width=0.85\columnwidth]{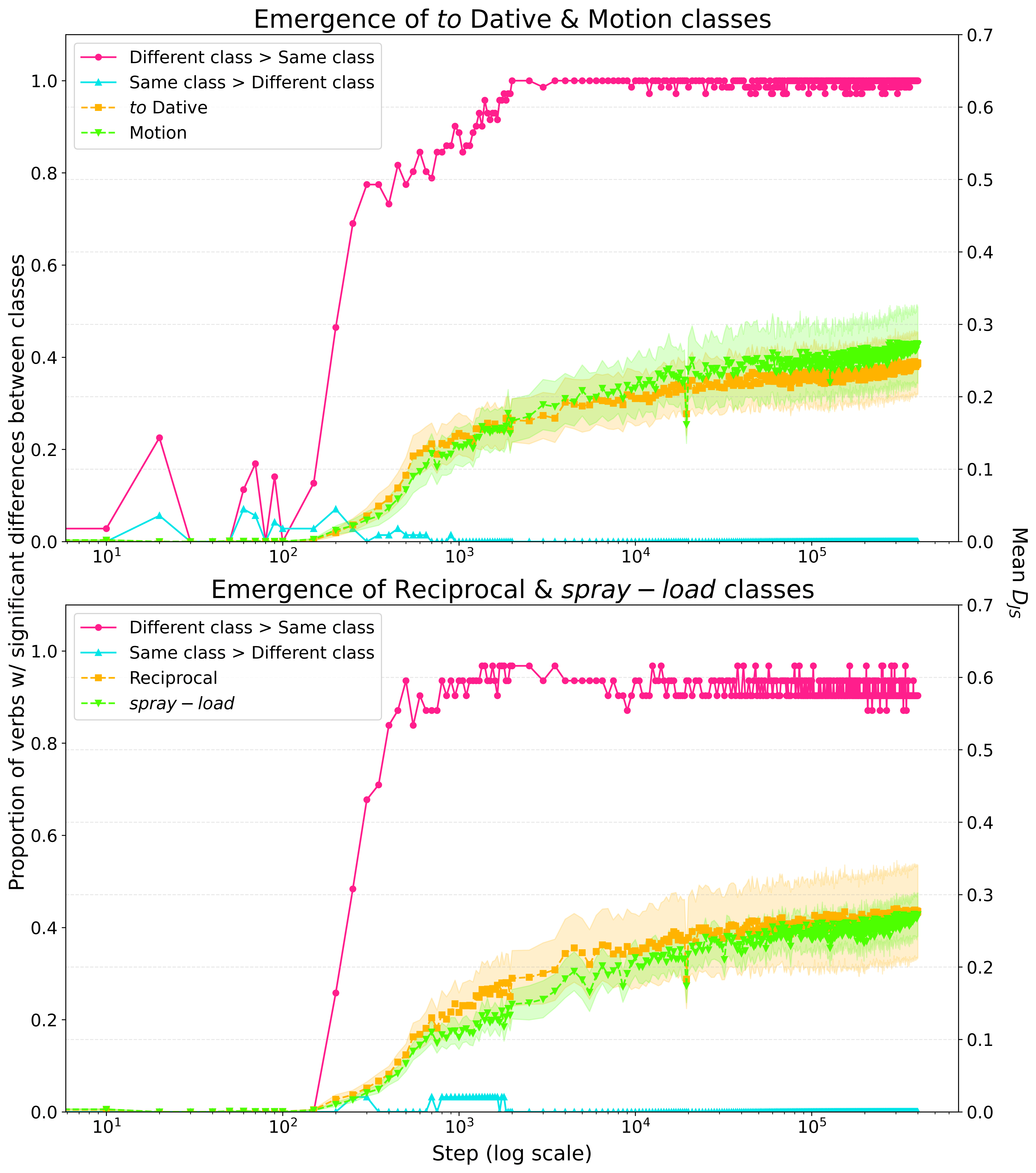}
    \caption{Mann-Whitney U-test between $U_A = \{D_{JS}(P_{v_{t}}(x) || P_v(x)) : v, v_{t} \in C_1\}$ (same class), $U_B = \{D_{JS}(P_{v_{t}}(x) || P_v(x)) : v_{t} \in C_1, v \in C_2\}$ (different class), for verb classes $C_1, C_2$. Mean of $D_{JS}$ across verbs in the same class. Shading is standard deviation.\vspace{-\baselineskip}}
    \label{fig:class-abstraction-plots}
\end{figure}

\paragraph{Item learning metric} Given pairwise $D_{JS}$ between all verbs across two classes, we measure the emergence of item-specific learning independent of effects of verb class as the mean $D_{JS}$ between verb \textit{v} and verbs within the same class. If in-class $D_{JS}$ increases, then idiosyncratic properties of verb \textit{v} that are not general to the class have been learned.\vspace{-0.5\baselineskip}
\paragraph{Class learning metric} Given $D_{JS}$ between verb \textit{v} with verbs from the same class $U_A$, and with verbs from the other class $U_B$ we use a one-tailed Mann-Whitney U-test to determine statistically significance \citep{mann1947test}. If mean $U_B$ is greater than $U_A$ and statistically significant under our test, then we can conclude that the model has learned \textit{some} systematic property across verbs belonging to a class that separates them from verbs belonging to the other class.

\subsection{Results: Distributional metrics}
In Figure \ref{fig:class-abstraction-plots}, we report the percentage of verbs for which the mean $D_{JS}$ with verbs from the opposite class are greater than the mean $D_{JS}$ with verb from the same class ($p < 0.001$ as a threshold). For the (i) \textit{to} Datives and (ii) Motion classes, class-based distinctions (pink) emerge 50 steps before in-class differences begin (green, orange), whereas for the (iii) Reciprocal and (iv) \textit{spray -- load} classes, the onset of change is concurrent. Verbs in neither class diverge from one another before there are statistically significant differences between the classes. Divergence between the two classes reaches its peak at different points in training -- step 3000 for the (i) \textit{to} Datives and (ii) Motion, step 1000 for (iii) Reciprocal and (iv) \textit{spray -- load} -- while verbs within the same class continue to diverge from one another until the end of the training. We note that there are a small number of verbs that are not significantly more similar to verbs in their canonical class (pink line does not always reach 1.0) than verbs in the comparison class, but that they are also not significantly more similar to verbs in the comparison class (blue line tends to 0).

\begin{figure}
    \centering
    \includegraphics[width=0.95\linewidth]{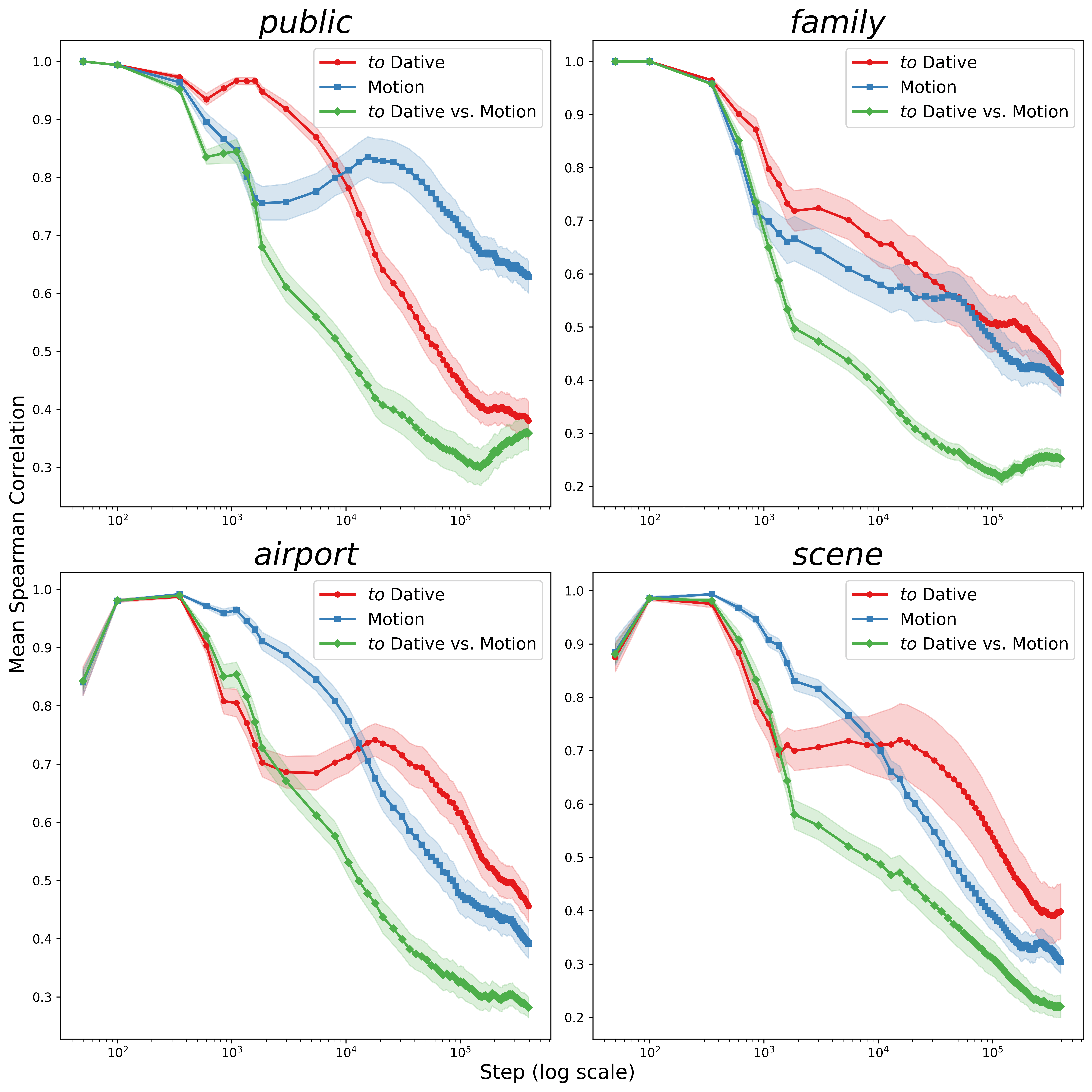}
    \caption{Mean pairwise Spearman correlations between prototype noun trajectories for each verb across training. Within-class and between-class means are reported with 95\% CI.\vspace{-\baselineskip}}
    \label{fig:noun-trajectories}
\end{figure}

Figure \ref{fig:jsd-class-plots} shows three early training steps before and after $D_{JS}$ begins to change. At Step 300, there is already clear separation of within-class (top-left, bottom-right) and out-of-class (top-right, bottom-left) quadrants when $D_{JS}$ starts increasing. This confirms that as soon as $D_{JS}$ increases, i.e., any change in verbal argument structure occurs, class-based distinctions are already evident.

\subsection{Results: Targeted behavioural evaluation}
We perform targeted evaluations with the larger \textit{to} Dative-Motion dataset. Four nouns assigned high probability by the fully trained model for the \textit{to} Dative and \textit{Motion} classes are tracked: \textit{public} and \textit{family}, and \textit{airport} and \textit{scene}, respectively.

For each noun and verb pair, we can derive a probability trajectory, i.e., what is the probability assigned to noun \textit{n} in contexts with verb \textit{v} at step \textit{t}. Computing Spearman correlations across these trajectories reveals how similar the relative change in probability assigned to \textit{n} is across verbs, within and across classes.

Trajectories for nouns which are prototypical arguments of a verb class remain highly correlated between verbs of that class earlier in training (\textit{public} and \textit{family} for \textit{to} Dative, \textit{airport} and \textit{scene} for Motion). Correlations of noun trajectories for verbs in different classes begin to decrease earlier in training, as well as correlations of noun trajectories for verbs in classes with which they are not prototypically found (\textit{airport} and \textit{scene} for \textit{to} Dative, \textit{public} and \textit{family} for Motion). Correlations between verbs in different classes are lowest. 

The sustained higher correlation across verbs for which a noun is prototypically an argument is consistent with the claim that general coarse-grained features, e.g., the semantic features associated with a broad verb class, are what is being learned early on in training (see Figure \ref{fig:noun-trajectories}). Lower correlations can be interpreted as verb-specific learning, since this means that the relative change in probability mass even differs across verbs in the same class. This does not occur until later in training for a noun's prototypical verb class. Lastly, correlations do not greatly \textit{increase} over the course of training suggesting that verb-specific learning does not precede class-general learning. These results further show that LMs have an abstraction-first bias.

\subsection{Exemplar-first baseline}
We confirm that the observed effects are not the result of an exemplar-first mechanism paired with properties of the training data by implementing a baseline for the exemplar-first mechanism. We use count-based vectors as a baseline since they obtain representations for a target verb as a function of the tokens it is observed with and their frequency. 
\begin{figure}[H]
    \centering
    \includegraphics[width=0.9\linewidth]{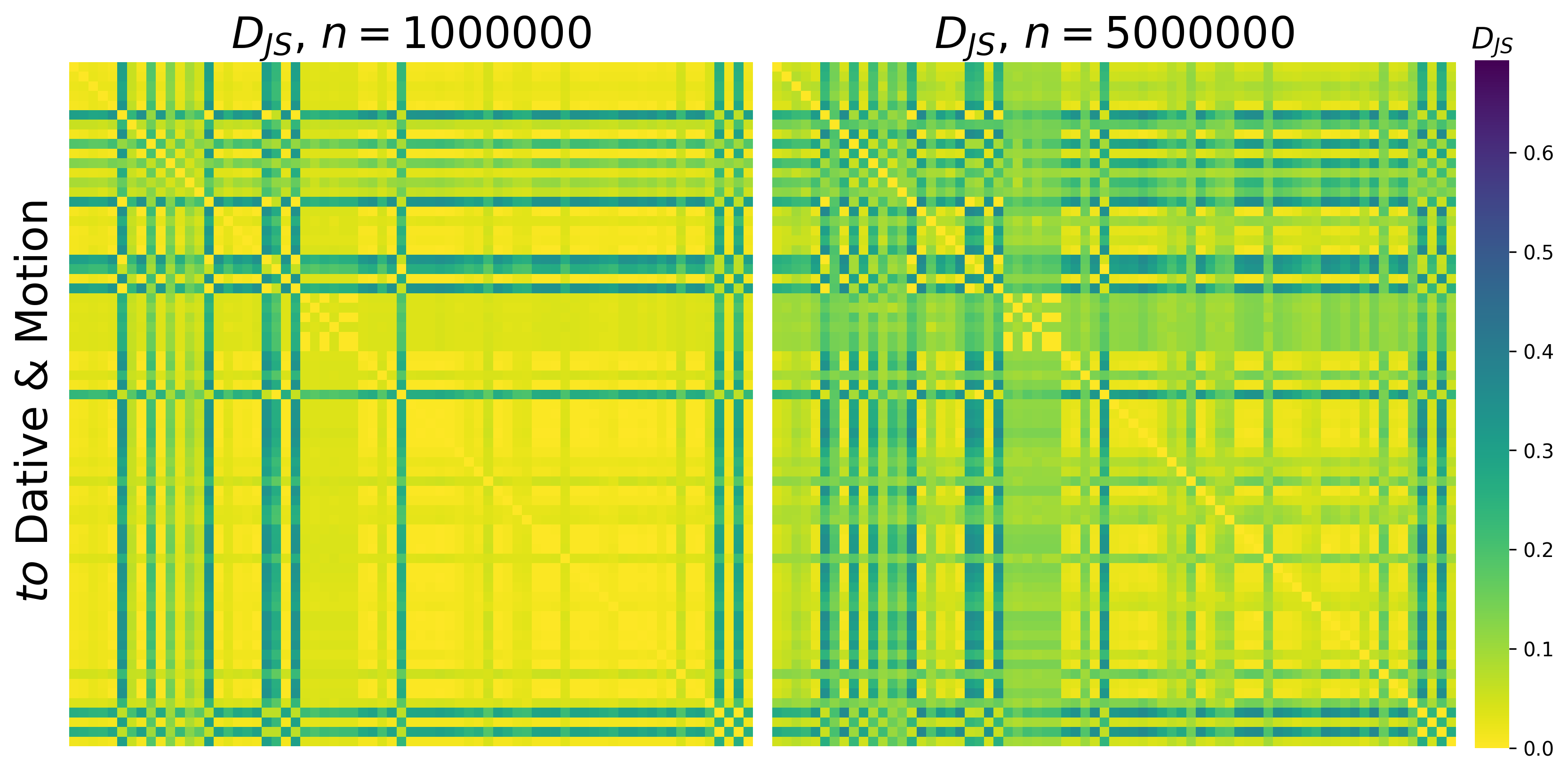}
    \includegraphics[width=0.9\linewidth]{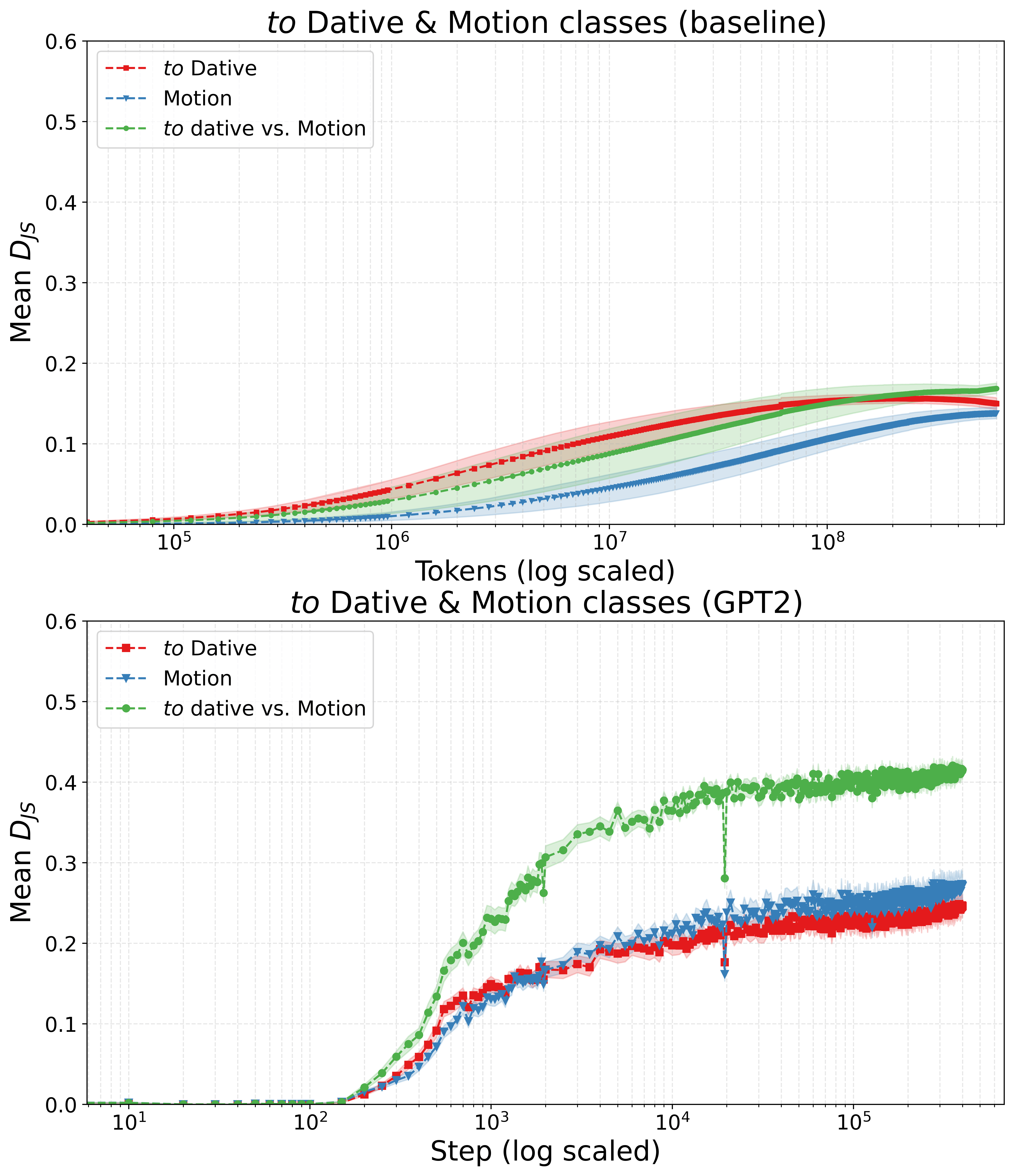}
    \caption{\textbf{Top:} Pairwise $D_{JS}$ across two classes in exemplar-first baseline show verb-specific patterns emerge without clear class distinctions.
    \textbf{Middle and bottom:} Exemplar-first baseline (middle) shows increasing divergence between verbs driven by verb-specific learning, before decrease, while GPT-2 is monotonic (bottom); baseline between-class mean divergences are not consistently greater than in-class means, unlike GPT-2. Shading is 95\% CI.\vspace{-\baselineskip}}
    \label{fig:baseline-results}
\end{figure}

We compute co-occurrence statistics for each verb in the correct prepositional structure, using the OpenWebText corpus tokenized with the GPT-2 tokenizer and a unidirectional 10 token context window. Stop word tokens like determiners were removed. Increasing the number of tokens observed is a proxy for training step. Add-$k$ smoothing ($k=0.5$) is applied to each vector and pairwise $D_{JS}$ is reported, as above. 

Results in Figure \ref{fig:baseline-results} are computed for $6.0 \times 10^7$ ($\sim$90\%) of the tokens containing the target verbs and structures. In the exemplar-first baseline, as hypothesized, verbs in the \textit{to} Dative class diverge from one another as a result of idiosyncratic patterns when less data is observed, before converging when more overlapping data is observed. Between-class distances are not greater than in-class distances until the \textit{to} Dative class converges, unlike with GPT-2 where these classes are differentiated early and consistently throughout training. This baseline confirms that observed LM learning trajectories do not match exemplar-first accounts.

\section{Experiment 2: Is learning generally abstraction-driven?}
GPT-2 learns argument structure classes using abstraction-first mechanisms. We investigate if this mechanism is broadly responsible for language learning using the two syntactic phenomena in Table \ref{tab:syntax-examples}: syntactic subcategorization (\textsc{Transitivity}) and filler-gap dependencies (\textsc{Relative Clause}). The same models are used.

\paragraph{Trajectories} If constructions are learned in tandem for classes of lexical items, then we can say that LMs are corresponding with an abstraction-first account. If they learn constructions idiosyncratically for individual instances, before gradually generalizing across classes, then LM learning better corresponds with an exemplar-first account.

\subsection{Data}
\begin{table}
    \centering
    \resizebox{\linewidth}{!}{
    \begin{tabular}{p{3.2cm} p{7.2cm}}
    \toprule
    \textsc{Structure} & \textsc{Example Sentences} \\
    \hline
    \multirow{2}{=}{\textsc{Transitivity} (\textit{n}=1000)} & The truck hadn't \textbf{astounded} \_\_ \\
                                         & The truck hadn't \textbf{fallen} \_\_ \\
    \hline
    \multirow{2}{=}{\textsc{relative clause} \\ (\textit{to} Dative; \textit{n}=824)} & The \textbf{person} that they sold the land to \_\_ \\
                                                                     & The person \textbf{thinks} that they sold the land to \_\_ \\
    \hline
    \multirow{2}{=}{\textsc{relative clause} \\ (Reciprocal; \textit{n}=288)} & The \textbf{person} that Optimus spoke with \_\_ \\
                                                              & The person \textbf{thinks} that Optimus spoke with \_\_ \\
    \bottomrule
\end{tabular}}
\caption{Example structures for Experiment 2. \textsc{Transitivity} is from BLiMP \citep{warstadt-etal-2020-blimp-benchmark}.\vspace{-\baselineskip}}
\label{tab:syntax-examples}
\end{table}

\paragraph{\textsc{Transitivity}} Verbs differ in whether or not they require an object: \textit{transitive}-\textit{intransitive} categorization. Transitive verbs require an object and intransitive verbs do not. Prediction distributions in a single frame across different verbs are used to track LMs learning of the category.

\begin{exe}
    \ex ... \textsc{verb} ... \textunderscore\textunderscore \label{ex:transitivity-frame}
\end{exe}

\paragraph{\textsc{Relative Clause}} Relative clauses are an example filler-gap dependency. In these structures, a noun that appears elsewhere in the sentence (the `filler') blocks a noun from surfacing locally to a verb which otherwise requires an object (the `gap'). These are structurally-conditioned categories (e.g., \textsc{verb\textsubscript{gap}} vs. \textsc{verb\textsubscript{no gap}}; \citealp{Gazdar1987, PollardSag1994HPSG, steedman+baldridge+2011}). We compare prediction distributions across two syntactic frames containing the same verb to track whether LMs have learned relative clause structures. We refer to (\ref{ex:rel-clause-1}) as the \textit{relative clause} environment and (\ref{ex:rel-clause-2}) as the \textit{matrix} environment.
\begin{exe}
    \ex \textit{The person that} ... \textsc{verb\textsubscript{gap}} ... \_\_ \label{ex:rel-clause-1}
    \ex \textit{The person thinks that} ... \textsc{verb\textsubscript{no gap}} ... \_\_ \label{ex:rel-clause-2}
\end{exe}

\paragraph{Dataset} We track learning here by comparing distributions between minimal pairs. For \textsc{Transitivity} we use minimal pairs from the BLiMP transitive test set \citep{warstadt-etal-2020-blimp-benchmark}. For \textsc{Relative Clause}, we templatically build minimal pairs using naturalistic data collected for Experiment 1. We use the \textit{to} Dative and Reciprocal classes since both permit human animate objects in the post-preposition position allowing us to create minimal pairs following the syntactic frames in (\ref{ex:rel-clause-1}) \& (\ref{ex:rel-clause-2}). Example data is provided in Table \ref{tab:syntax-examples}.
\begin{figure}
    \centering
    \includegraphics[width=0.8\linewidth]{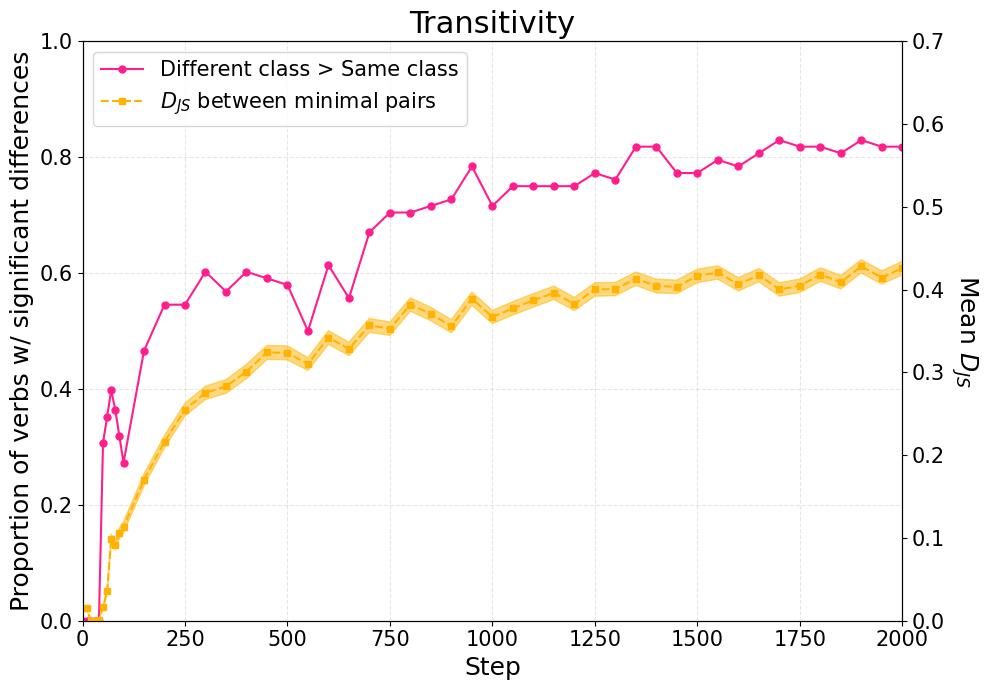}
    \caption{Comparing class-level and item-based behaviour for \textsc{Transitivity} dataset.}
    \label{fig:transitivity}
\end{figure}

\subsection{Setup}
The same paradigm discussed in Section \ref{sec:paradigm} is used here, taking $D_{JS}$ over next-token prediction distributions at the target position given two prefixes containing the structures being compared.

\paragraph{Item learning metric} Given pairwise $D_{JS}$ between distributions predicted for a single minimal pair, we measure the emergence of item-specific learning independent of effects of abstractions, e.g., an abstract transitive class or a gap class, as the mean $D_{JS}$ across all minimal pairs in a single condition. If in-class $D_{JS}$ increases, then learning has occurred for single examples that are not tied to learning a generalized construction.

\paragraph{Class learning metric} We define class learning here again by taking a mean across the distributions conditioned on prefixes containing the same verb. We take pairwise $D_{JS}$ between all the distributions for each verb and perform a one-tailed Mann-Whitney U-test to determine the proportion of verbs that are significantly more similar to verbs in the same category than to verbs from the opposite category. For \textsc{Transitivity} categorization, the verbs in the two categories are non-overlapping. For the \textsc{Relative Clause} categorization, the two categories contain the same verbs in different syntactic contexts, e.g., \{\textit{give\textsubscript{no gap}}, \textit{sell\textsubscript{no gap}}, \textit{grant\textsubscript{no gap}}\} and \{\textit{give\textsubscript{gap}}, \textit{sell\textsubscript{gap}}, \textit{grant\textsubscript{gap}}\}. As before, this measure only increases if LMs treat verbs in the same syntactic category similarly.
\begin{figure}
    \centering
    \includegraphics[width=0.8\linewidth]{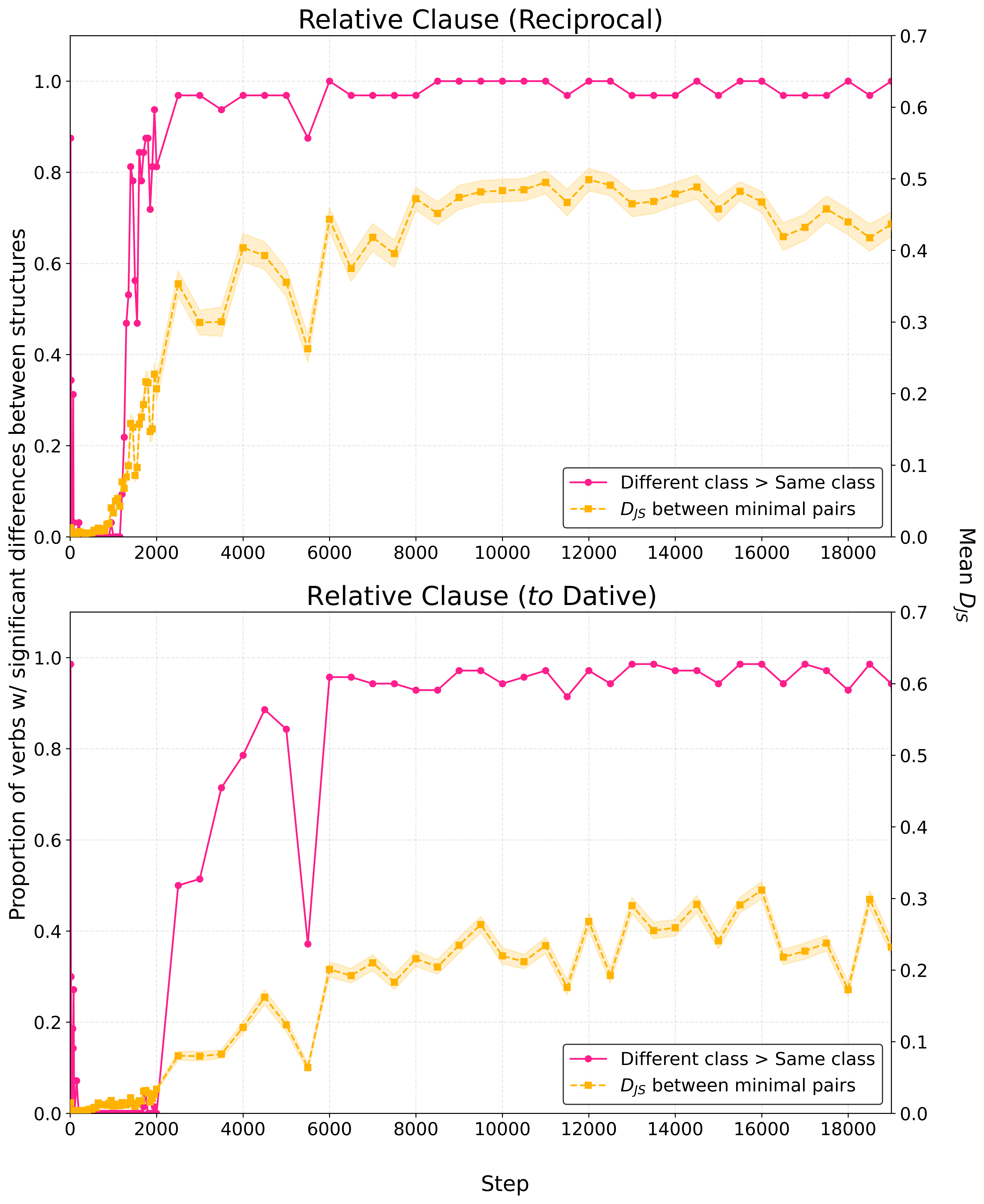}
    \caption{Comparing class-level and item-based behaviour for \textsc{Relative Clause} datasets.}
    \label{fig:rel-clauses}
\end{figure}

\subsection{Results}
\paragraph{GPT-2 consistently learns via abstraction.} Across all structures tested, class-general behaviour emerges as soon as any item-based behaviour is evident. Figures \ref{fig:transitivity} and \ref{fig:rel-clauses} show that an increase in pairwise $D_{JS}$ between minimal pairs (orange), always occurs with increases in verbs which are systematically more similar to those belonging to the same category, than those which do not (pink). Class-like behaviour does not emerge slowly -- learning a construction always corresponds with generalizing that construction across a class of items. This matches our findings for the learning of semantic verb argument structure classes.

An unexpected separation into categories at the beginning of training is observed in the relative clause conditions. We confirm that this is due to the token \textit{thinks} appearing in all non-relative clause sentences by varying between nine embedding verbs for each target verb. This reduces the effect (see Appendix \ref{sec:embedding-verbs}, Figure \ref{fig:varied-embedding-verbs}).

\begin{figure}
    \centering
    \includegraphics[width=\linewidth]{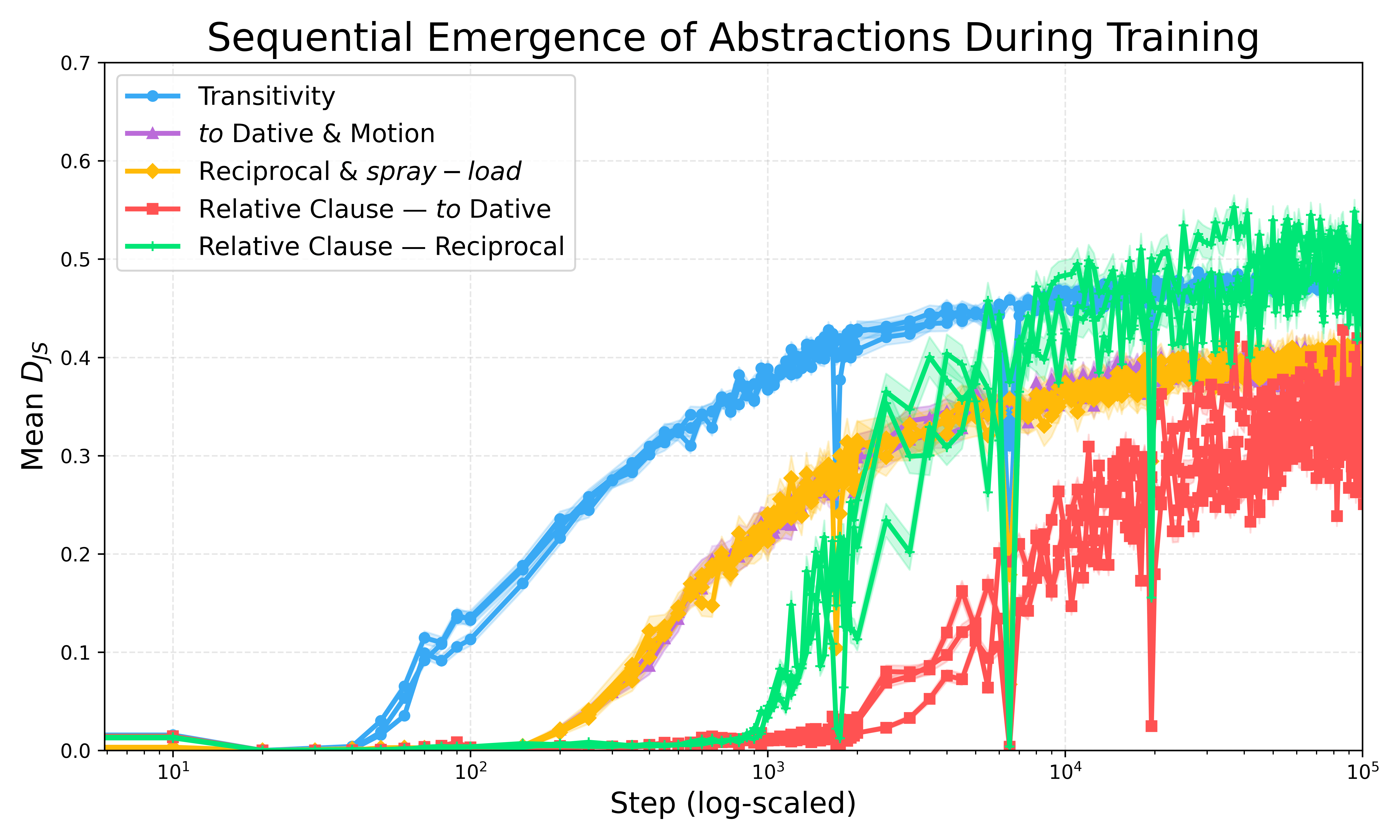}
    \caption{Learning curves for each abstraction plotted for three models trained with different random seeds.}
    \label{fig:combined-over-training}
\end{figure}

\paragraph{Structures are not learned globally.} Relative clause filler-gap categories are not learned for all verbs at once. Rather, they are learned separately for verb classes: pairwise $D_{JS}$ increases for relative clauses containing Reciprocal verbs earlier than for relative clauses containing \textit{to} Dative verbs. This is confirmed by comparing breakpoints for each verb: the median across verbs for the Reciprocal class and the \textit{to} Dative class are 950 and 1900, respectively, matching the onsets observed in Figure \ref{fig:rel-clauses}. An unpaired t-test confirms that the differences in breakpoints for the two classes are statistically significant ($p < 10^{-8}$). Breakpoints are heuristically computed as the step after which the average $D_{JS}$ for a verb across minimal pairs is consistently at least 0.01 greater than the average of the first 30 steps of training.

\paragraph{Abstractions are learned sequentially.} Across three runs, we see that each of the abstractions emerges with distinct onsets in the same sequence over the course of training (see Figure \ref{fig:combined-over-training}). Before the onset of learning for a given construction, mean $D_{JS}$ in each condition does not increase and is around 0. This confirms further that gradual memorization does not drive syntactic learning.

\section{Discussion}
Sequential emergence of linguistic behaviour cannot be due to shifts in the training data or objective, as this is uniform over standard LM pretraining (see discussion in \citealp{chang-etal-2024-characterizing}). It must be the case that behavioural change is due to shifts in model representations or mechanisms. We have shown here that these shifts are best viewed as resulting from learning \textit{abstractions}, rather than storage of \textit{exemplars}. Converging evidence comes from \citet{arora-etal-2024-causalgym}, who show that even \textit{individual} phenomena involve discrete learning stages. Our results are also consistent with findings from \citet{wilson-etal-2023-abstract}, who show that LMs can learn abstractions which hold over classes of lexical items by the end of training, the types of abstractions we test here. \citeauthor{wilson-etal-2023-abstract} further discuss abstractions that are difficult for LMs to learn -- future work into the learning trajectories of these phenomena may provide further insight into why LMs are unable to learn them.

The order that abstractions emerge in is consistent across training runs: \textit{syntactic} subcategorization (\textsc{transitivity}, \textit{t} < 100), \textit{semantic} argument structure properties (\textsc{verb classes}, \textit{t} > 100), lastly, non-local dependencies (\textsc{relative clause}, \textit{t} > 1000). Our results differ from \citet{evanson-etal-2023-language}, who find a similar ordering, but report gradual parallel learning, while we find clear evidence of sequential emergence, though our experimental setup differs as it directly measures argument predictions, rather than using subject-verb agreement as a proxy. Discrete learning stages provide direct support for abstraction-first views of acquisition -- the gradual emergence expected under exemplar-first acquisition is not empirically observed. Abstraction-first theories include those which propose that learning complex structures requires simpler abstractions to be in place \citep{lebeaux2000language, friedmann2021growing, Diercks2023DMS}; the acquisition order we observe is consistent with the predictions of these theories, as well. This being said, acquisition order may in principle also arise from differences in construction frequency, and, as such, cannot be itself an argument for complexity-based theories. The role of frequency effects and abstraction--driven learning biases in determining acquisition order must be teased apart in future work.

Despite apparent parallels, a gap between many theories of child language acquisition and the theory of LM learning we have presented here is that all abstract linguistic categories learned by LMs are a product of the data observed during training and their inductive biases. The abstraction-first behaviour observed is not the result of predetermined linguistic features encoded in the model before training, unlike what has been proposed to be the case for humans with, e.g., the mapping between argument positions and semantic roles \citep{Pinker1989learnability}. Indeed, earlier work has shown that abstractions like argument structure classes are learnable from distributional patterns, albeit with some supervision, \citep[e.g.,][]{lapata-1999-acquiring, schulte-im-walde-2000-clustering}.

\section{Conclusion}
A leading view on how LMs can contribute to our knowledge of human linguistic cognition is as \textit{existence proofs}: the mechanisms LMs employ to acquire the constructions that they do are \textit{in principle} possible ways that human learners may arrive at knowledge of those same constructions \citep{portelance2024roles}. As such, what we understand to be the inductive biases of LMs changes the conclusions we can draw on the basis of their success. 

Some have recently discussed that LMs may indeed be an existence proof for exemplar-based models \citep{Ambridge2020Against, ambridge2024large, goldberg2024usage}. For example, \citeauthor{Ambridge2020Against} suggests modern LMs are compatible with a view where linguistic abilities arise via exemplar-storage because the high parameter count of these models in principle allows them to memorize. Following a body of literature that has cautioned against taking connectionist models as inherently undermining symbolic views like the abstraction-first account of learning discussed here \citep{Fodor+Pylyshyn+1988, newmeyer2003grammar, mahowald2020counts}, we consider this to be a testable empirical question. Indeed, our results show that the abstraction-first account is a viable explanation of LM learning. An open question for future work is characterizing the sorts of abstractions that LMs can learn -- are LM abstractions best understood at the word-level or a more dynamic construction-level? Evaluating LM behaviour over the course of training may be a fruitful direction for answering questions like these and uncovering more mechanisms underlying their success.

\section*{Limitations}
Our study is limited by its focus on a single model architecture trained and evaluated on a single language. This choice was motivated by previous uses of these models in similar studies (see Section \ref{sec:data-and-models}), and constrained by the resource requirements associated with training and storing model checkpoints. Our dataset can be applied to a wider-range of English models and the paradigm across languages. These studies will help us understand how general the effects we report here are. Further testing on different models may help localise whether this is an effect of the model architecture and training procedure and testing on models of different sizes can answer whether parameter count has an effect as well.

The other limitation that we identify here is the breadth of the linguistic phenomena studied. While we have focused on abstraction within the verbal domain, abstraction occurs at different levels of linguistic representation, including below the word level. Derivational suffixes like English \textit{-ness} which can combine with a range of adjectives to derive nouns is another example of abstraction -- here, over syntactic categories. Whether LMs can productively combine and use such suffixes and whether this displays the same abstraction-first learning that we observe here would be an interesting avenue for future work.

Lastly, while we have provided converging \textit{behavioural} evidence in favour of abstraction-first learning, understanding how \textit{representations} change to produce these behaviours is a core desideratum. We hope that our datasets may be helpful for future studies along these lines. 

\section*{Acknowledgements}
The authors would like to thank the three anonymous reviewers for constructive feedback, as well as Dan Jurafsky, Alex Warstadt, Beth Levin, Boris Harizanov, and members of the Stanford NLP group for much discussion. JJ is supported in part by funding from the Social Sciences and Humanities Research Council of Canada.

\bibliography{custom, anthology}

\appendix

\section{Datasets: \textsc{Argument Structure Classes}}
Verbs used for each argument structure class:

\paragraph{\textit{to} Dative} \textit{sell} (85), \textit{grant} (85), \textit{issue} (84), \textit{explain} (81), \textit{lease} (79), \textit{award} (79), \textit{deny} (79), \textit{teach} (75), \textit{present} (75), \textit{offer} (75), \textit{allocate} (74), \textit{give} (66), \textit{assign} (65), \textit{pass} (65), \textit{read} (63), \textit{relay} (62), \textit{show} (62), \textit{cede} (62), \textit{pay} (56), \textit{trade} (52), \textit{deliver} (51), \textit{feed} (49), \textit{rent} (48), \textit{transfer} (43), \textit{forward} (33), \textit{promise} (30), \textit{mail} (28), \textit{allot} (27), \textit{tell} (26), \textit{guarantee} (24), \textit{yield} (18), \textit{dictate} (16), \textit{ask} (12), \textit{bequeath} (7), \textit{refund} (5). (\citealp{levin1993english}; p.45-46)

\paragraph{Motion} \textit{go} (130), \textit{travel} (124), \textit{walk} (119), \textit{fly} (116), \textit{flee} (114), \textit{sail} (111), \textit{rush} (108), \textit{descend} (95), \textit{cross} (94), \textit{run} (93), \textit{march} (93), \textit{climb} (83), \textit{swim} (77), \textit{ascend} (76), \textit{drift} (70), \textit{ride} (65), \textit{arrive} (56), \textit{hurry} (49), \textit{race} (36), \textit{float} (32), \textit{journey} (29), \textit{depart} (19), \textit{crawl} (16), \textit{row} (15), \textit{taxi} (14), \textit{hasten} (12), \textit{limp} (11), \textit{chase} (11), \textit{glide} (11), \textit{hike} (10), \textit{dash} (10), \textit{wander} (10), \textit{scramble} (9), \textit{cycle} (8), \textit{tumble} (8), \textit{meander} (8). (\citealp{levin1993english};  p.263-269)

\paragraph{Reciprocal} \textit{collaborate} (48), \textit{team} (47), \textit{talk} (46), \textit{mate} (44), \textit{communicate} (41), \textit{compete} (39), \textit{clash} (35), \textit{join} (35), \textit{meet} (30), \textit{cooperate} (28), \textit{speak} (19), \textit{correspond} (16), \textit{fight} (15), \textit{connect} (10), \textit{agree} (8), \textit{play} (4). (\citealp{levin1993english}; p.59)

\paragraph{\textit{spray}-\textit{load}} \textit{inject} (93), \textit{load} (66), \textit{stock} (63), \textit{plant} (31), \textit{spray} (31), \textit{stuff} (22), \textit{pack} (20), \textit{hang} (15), \textit{smear} (12), \textit{rub} (11), \textit{plaster} (10), \textit{shower} (10), \textit{scatter} (7), \textit{sprinkle} (7), \textit{drape} (6).  (\citealp{levin1993english}; p.50)

Each class is uniform with respect to a broad categorization of argument preferences structure. Future work may investigate how finer-grained distinctions that we do not assess here, e.g., participation in argument structure alternations, are learned over the course of training.

\section{Datasets: \textsc{Relative Clause}}
The \textsc{Relative Clause} datasets were automatically generated using the sentences collected for each verb class for Experiment 1. Experiment 1 does not restrict sentences collected to being a full finite clause with a subject and active voice verb, so Stanza parses are used to extract the subject (and the direct object of the verb for \textit{to} Dative class). Sentences for which the sequence of words extracted for the subject and object were not contiguous, or for which one of the required arguments could {not} be extracted on the basis of the parse tree, were discarded, accounting for the decrease in the number of sentences between the full dataset and the derived relative clause dataset.

Verbs were converted to a simple past form and placed in two different syntactic templates with its extracted subject and object, corresponding to the relative clause (\ref{ex:rel-clause-template-1})-(\ref{ex:rel-clause-template-2}) and non-relative clause (\ref{ex:matrix-template-1})-(\ref{ex:matrix-template-2}) structures being compared in this condition.

\begin{exe}
    \ex The person that \textsc{subject} V-\textsc{ed} \textsc{obj} to \_\_. \label{ex:rel-clause-template-1}
    \ex The person that \textsc{subject} V-\textsc{ed} with \_\_.\label{ex:rel-clause-template-2}
    \ex The person thinks that \textsc{subject} V-\textsc{ed} \textsc{obj} to \_\_. \label{ex:matrix-template-1}
    \ex The person thinks that \textsc{subject} V-\textsc{ed} with \_\_.\label{ex:matrix-template-2}
\end{exe}

\subsection{Alternating embedding verbs}\label{sec:embedding-verbs}
\begin{figure}
    \centering
    \includegraphics[width=0.8\linewidth]{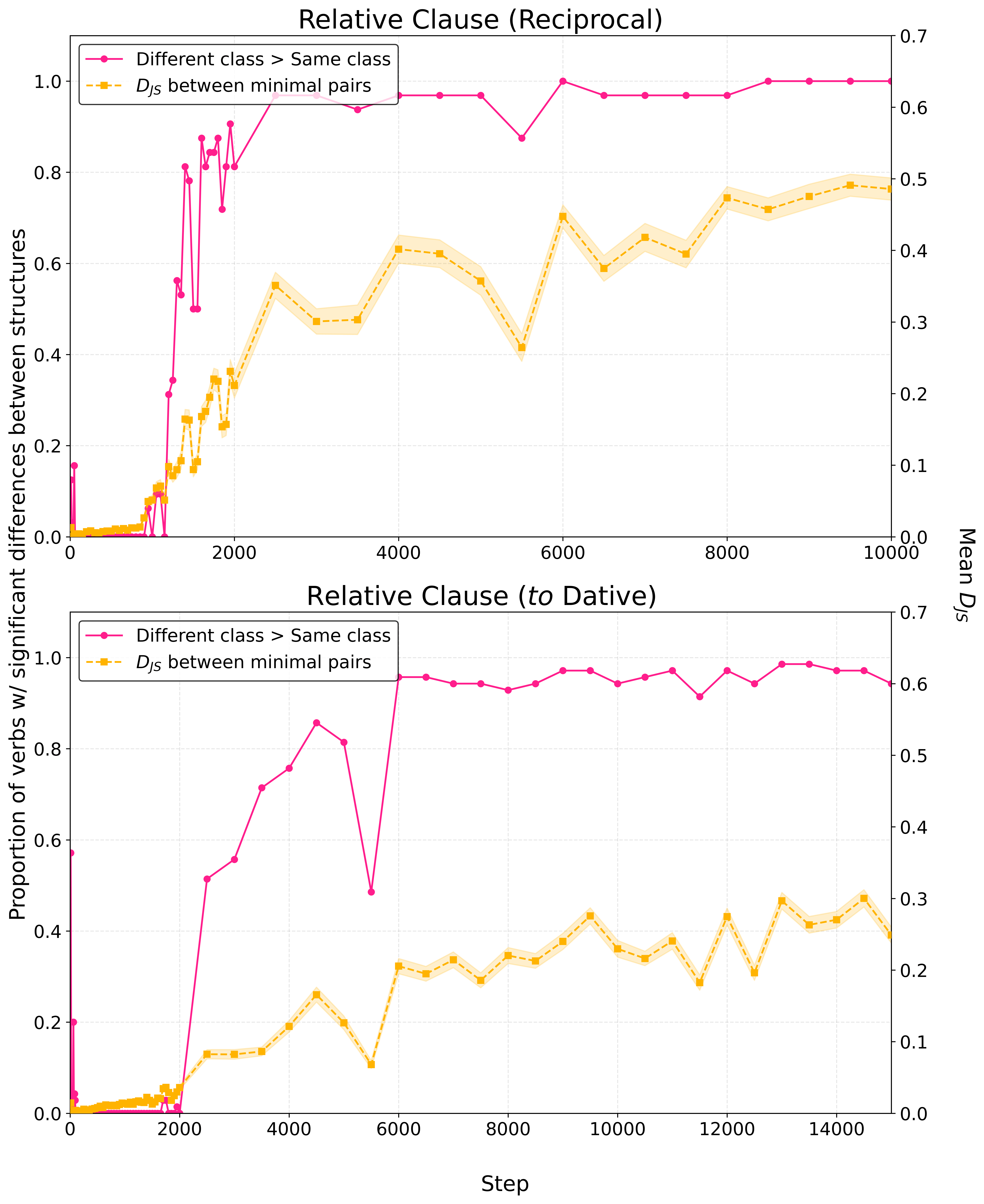}
    \caption{Comparing class-level and item-based behaviour on \textsc{Relative Clause} datasets with varied embedding verbs. \vspace{-\baselineskip}}
    \label{fig:varied-embedding-verbs}
\end{figure}
In order to construct minimal pairs with the relative clause condition that should not license a filler-gap dependency that contain the relativized noun \textit{the person} and the complementizer \textit{that}, we embed target sentences under the verb \textit{think}, as shown above. However, this introduces an additional word \textit{not} present in the relative clause condition, the embedding verb \textit{think} itself. This can cause an unexpected differentiation between conditions early on in training (see Figure \ref{fig:rel-clauses}). We confirm that this is the effect of the presence of the word \textit{think} itself and not early syntactic learning by varying the embedding verbs: \textit{thinks}, \textit{believes}, \textit{knows}, \textit{says}, \textit{claims}, \textit{announces}, \textit{states}, \textit{reports}, \textit{reveals}. They occupy the position of \textit{thinks} in (\ref{ex:matrix-template-1}) \& (\ref{ex:matrix-template-2}). Figure \ref{fig:varied-embedding-verbs} shows that the spike observed in Figure \ref{fig:rel-clauses} at early steps is no longer present.

\section{Compute, Packages, and Licenses}
Experiments reported used NLTK (3.6.4), lemminflect (0.2.3), and Stanza (1.8.2) for preprocessing and parsing during the creation of the dataset; transformers (4.40.1), scipy (1.3.1), numpy (1.18.2) and SpaCy (2.1.8) were used for computational experiments. All experiments were run on NVIDIA RTX or NVIDIA RTX A6000 GPUs depending on cluster availability.

All data (WikiText-103, \citealp{merity2016pointer}; BLiMP, \citealp{warstadt-etal-2020-blimp-benchmark}) and pretrained models (GPT-2 checkpoints, \citealp{karamcheti2021mistral}) are freely available for download and were used standardly. Our dataset and code can be found at \texttt{https://github.com/jasperjjian/abstraction}.
\end{document}